%% file: main.tex
\documentclass[conference]{IEEEtran}
\IEEEoverridecommandlockouts
\usepackage{cite}
\usepackage{amsmath,amssymb,amsfonts}
\usepackage{algorithm}
\usepackage{algpseudocode}
\usepackage{graphicx}
\usepackage{textcomp}
\usepackage{xcolor}
\def\BibTeX{{\rm B\kern-.05em{\sc i\kern-.025em b}\kern-.08em
    T\kern-.1667em\lower.7ex\hbox{E}\kern-.125emX}}

\usepackage{hyperref}
\hypersetup{
    colorlinks=true,
    linkcolor=blue,
    citecolor=blue,      
    urlcolor=blue,
}
\usepackage{libertine}
\usepackage{cleveref}
\usepackage{todonotes}
\usepackage{subcaption}
\graphicspath{{images/}}
\usepackage{csquotes}
\usepackage{bbm}
\usepackage{multirow}

\begin{document}

\title{Surrogate Infeasible Fitness Acquirement FI-2Pop for Procedural Content Generation}


\graphicspath{{images/}}

\author{
    \IEEEauthorblockN{Roberto Gallotta}
    \IEEEauthorblockA{\textit{Araya Inc.} \\ Tokyo, Japan}
    \and
    \IEEEauthorblockN{Kai Arulkumaran}
    \IEEEauthorblockA{\textit{Araya Inc.} \\ Tokyo, Japan}
    \and
    \IEEEauthorblockN{L. B. Soros}
    \IEEEauthorblockA{\textit{Cross Labs} \\ Kyoto, Japan}
}

\maketitle

\input{sections/00_abstract}


\input{sections/01_introduction}

\input{sections/02_background}

\input{sections/03_domain}

\input{sections/04_proposed_variant}

\input{sections/05_results}

\input{sections/06_discussion}

\input{sections/07_acks}

\bibliographystyle{ieeetr}
\bibliography{references}

\end{document}

%% file: sections/00_abstract.tex
\begin{abstract}
    When generating content for video games using procedural content generation (PCG), the goal is to create functional assets of high quality. Prior work has commonly leveraged the feasible-infeasible two-population (FI-2Pop) constrained optimisation algorithm for PCG, sometimes in combination with the multi-dimensional archive of phenotypic-elites (MAP-Elites) algorithm for finding a set of diverse solutions. However, the fitness function for the infeasible population only takes into account the number of constraints violated. In this paper we present a variant of FI-2Pop in which a surrogate model is trained to predict the fitness of feasible children from infeasible parents, weighted by the probability of producing feasible children. This drives selection towards higher-fitness, feasible solutions. We demonstrate our method on the task of generating spaceships for Space Engineers, showing improvements over both standard FI-2Pop, and the more recent multi-emitter constrained MAP-Elites algorithm.
\end{abstract}

%% file: sections/01_introduction.tex
\section{Introduction}
    Designing content is still one of the most time-consuming steps in developing video games. Many games nowadays include some kind of procedural content generation (PCG) to reduce the time spent in developing content, be it levels, interactive items or even entire quest lines \cite{hendrikx_procedural_2013}. One usual PCG approach is to search within a space of possible content for solutions with good quality. This search process can be seen as an optimisation problem, where the quality of the content is defined by an utility function. In some cases there are restrictions imposed upon the content, which can be expressed as constraints in the optimisation problem. The feasible-infeasible two-population (FI-2Pop) evolutionary algorithm \cite{kimbrough_feasibleinfeasible_2008} has been widely used in constrained optimisation tasks due to its simplicity and intuitiveness. FI-2Pop has been used for PCG extensively, as it allows the programmer to specify domain-specific constraints and fitness functions \cite{baldwin_mixed-initiative_2017}.
    
    FI-2Pop is able to freely explore the solution space in its entirety, regardless of feasibility (constraint satisfaction), so that feasible solutions that would otherwise be unreachable due to the feasibility region's topology can be reached via infeasible solutions. In the original algorithm, infeasible solutions are selected for evolution in inverse proportion to the number of constraints broken, but this selection mechanism does not directly optimise for feasible solution fitness. In this work we propose a novel variant of FI-2Pop, training a surrogate fitness function that predicts the fitness of infeasible solutions' offspring, weighted by the probability of generating a feasible child, which can accelerate the optimisation process. We apply our algorithm to PCG for creating novel spaceships in the video game Space Engineers, improving upon prior work based on standard FI-2Pop \cite{gallotta_evolving_2022}. We then show our offspring fitness prediction method's effectiveness in a mixed-initiative setting, solving the spaceship generation task with the constrained multi-dimensional archive of phenotypic-elites (CMAP-Elites) algorithm \cite{khalifa_talakat_2018}. Finally, we extend the latter application to include an optimising emitter \cite{fontaine_covariance_2020}, which uses our novel infeasible fitness function to balance both fitness and coverage of solutions.

%% file: sections/02_background.tex
\section{Background}\label{sec:background}
    \subsection{Procedural Content Generation}\label{sec:background_pcg}
        PCG is the process of algorithmically creating components for an application, either independently or in tandem with a human designer. PCG has been widely used in video games, creating everything from assets to entire levels \cite{hendrikx_procedural_2013}.
        
        Evolutionary algorithms (EAs) are a family of population-based search algorithms well-suited for PCG: by assigning a fitness value to solutions, EAs can be used to search for desirable content. To do so while retaining a wide set of solutions to be chosen from, one can use EA-based \enquote{quality-diversity} algorithms, for example, MAP-Elites \cite{mouret_illuminating_2015,gravina_procedural_2019}.
        
    \subsection{FI-2Pop}\label{sec:background_fi2pop}
        The FI-2Pop \cite{kimbrough_introducing_2004, kimbrough_feasibleinfeasible_2008} EA evolves two separate populations of solutions to follow a fitness signal in a constrained domain. One population is comprised solely of feasible solutions and the other solely of infeasible solutions. Each feasible solution has no constraint violations and its fitness is determined by a domain-specific utility function, whereas each infeasible solution violates at least one constraint and its fitness is the inverse of the number of constraints violated.
        
        The EA loop of FI-2Pop is as follows: one or more solutions are picked from each population (based on their fitness) as parents to generate new offspring solutions via crossover and mutation. Each parent may generate either a feasible or infeasible offspring, regardless of its own feasibility. All offspring are then added back to the populations based on their feasibility. If the size of a population exceeds a set limit, solutions are removed from it based on their fitness. This loop is then repeated for a given number of generations.
        
        Selecting infeasible solutions inversely to the number of constraints violated makes selection pressure towards feasibility a generic repair mechanism. While simple and effective, this heuristic could be replaced by a more information-rich function for optimisation, as we propose in \cref{sec:proposed_variant}.
        
    \subsection{Surrogate-Assisted Evolutionary Algorithms}\label{sec:background_saeas}
        Surrogate-assisted EAs (SAEAs) \cite{tong_surrogate_2021} use a surrogate model to approximate the fitness function in computationally-expensive problems. SAEAs use a surrogate model that is trained on the fitness of solutions collected from previous runs. The model then interacts with the EA by estimating the fitness of newly-generated solutions. SAEAs typically include a re-evaluation step where some or all solutions in the current generation are evaluated with the true fitness function, to be used in subsequent training of the model. This online training helps ensure convergence of the EA by improving the modelling of relevant areas of the search space.
        
        There are many classes of fitness models; in our work we use a regression-based absolute fitness model, where the estimator takes features of the solution and is trained to predict the fitness directly \cite{tong_surrogate_2021}. However, in contrast to most uses of SAEAs, we use our model to create a different fitness function than the original algorithm's (FI-2Pop's). Our proposed fitness function is similar to \enquote{fitness inheritance} \cite{smith_fitness_1995,chellapilla_fitness_1999}, where the estimation is based on the parents' fitness, although we assign fitness based on the fitness of the parents' \emph{prior offspring}. We call this novel estimation \enquote{acquirement}, and give more details in \cref{sec:proposed_variant}.
        
    \subsection{MAP-Elites}\label{sec:background_mapelites}
        MAP-Elites \cite{mouret_illuminating_2015} is an \enquote{illumination algorithm} that searches for a diverse set of high-performing solutions, as defined by behaviour characteristics (BCs) of interest. MAP-Elites projects and subdivides the search space into BC-based niche bins, each containing one or more solutions. In standard MAP-Elites, the underlying search process is unguided and unbiased: at each step, a random bin is selected amongst all non-empty bins with uniform probability. New solutions are then generated from the selected bin and then assigned to their respective bins. Bins have a fixed capacity, so new solutions can take the place of older or less-fit solutions.
        
        MAP-Elites requires an optimisation algorithm in order to produce new solutions, and EAs are typically used for this in practice. For PCG with constraints, the most common choice is FI-2Pop. There are multiple variations on MAP-Elites, such as letting a bin contain more than one solution (elite), or allowing near-bin crossover. Other variations have focused on the bin selection process itself: \textit{emitters} \cite{fontaine_covariance_2020} can be used to more directly optimise for average fitness or coverage. Recent work \cite{cully_multi-emitter_2021} extended this to a meta-optimisation process, whereby a bandit algorithm selects between different types of emitters to optimise for given metrics. MAP-Elites has been augmented with surrogate models before, using Gaussian processes \cite{gaier_data-efficient_2017} and neural networks \cite{zhang_dsame_2022}; however, these simply model the original fitness function instead of replacing it. In \Cref{sec:results} we demonstrate FI-2Pop with our novel fitness function extended to the CMAP-Elites setting, and with the use of emitters.

%% file: sections/03_domain.tex
\section{Domain}\label{sec:domain}
    Space Engineers is a 3D sandbox game with realistic graphics and physics, played by approximately 5,000 players daily, where the objective is to build spaceships and other structures to collect resources, navigate through space, and interact with other characters. Structures are formed out of interacting blocks, each with their own properties and functionalities. In order to be usable, these structures must satisfy various functional constraints. For example, a spaceship needs a cockpit, an engine, and thrusters.
    
    Procedurally generating structures for Space Engineers involves creating solutions that both respect constraints defined by the game itself, as well as any additional, hand-engineered hard or soft constraints, whilst simultaneously striving to find aesthetically-pleasing results. We follow previous work \cite{gallotta_evolving_2022}, which introduced a hybrid EA consisting of multiple modular L-systems \cite{lindenmayer_mathematical_1968} and FI-2Pop, to combine domain knowledge with constrained optimisation in order to procedurally generate spaceships. We use the same feasible fitness function and constraints. The hard constraints are, firstly, no intersections between blocks, and secondly, the structure must contain all required blocks. The soft constraint is to have symmetry along any axis. The feasible fitness function is based on a probability density estimate of various metrics, derived from human-designed spaceships made available on Steam Workshop. The 4 metrics used are the ratio between the number of functional blocks and the total number of blocks, the ratio between the filled and total volume, the ratio between the major and medium axis of the ship, and the ratio between the major and smallest axis.

%% file: sections/04_proposed_variant.tex
\section{Surrogate Infeasible Fitness Acquirement}\label{sec:proposed_variant}
    As discussed previously, the infeasible fitness for standard FI-2Pop selects infeasible parents likely to result in feasible offspring (by virtue of having relatively few constraint violations), but this fitness calculation does not account for offspring fitness. We propose using a surrogate model to estimate the fitness of infeasible parents' children, weighted by the probability of the children being feasible, driving selection towards high-performing, feasible solutions. This can be seen as encouraging evolvability \cite{altenberg_evolution_1994} in the infeasible population, where parents are chosen directly based on their probability of generating children with higher fitnesses.
    
    The main FI-2Pop loop involves selecting parents to create children using crossover and mutation, after which children are placed into their respective populations. Selection and evolution happen separately for the feasible and infeasible populations. In our variant, we keep a buffer for each infeasible parent that contains the fitnesses of any feasible children it produces; we also keep a tally of the empirical probability of the infeasible parent producing a feasible child. We then train a surrogate model\textemdash in our experiments, a fully-connected neural network\textemdash to directly predict a statistic of the children's fitness, weighted by the probability of producing a feasible child, using the mean-squared objective.
    
    In preliminary experiments, we tested different statistics of the children’s fitness, including the mean (denoted \enquote{$\mu$}), the maximum (denoted \enquote{M}), the minimum (denoted \enquote{m}), as well as the upper and lower confidence bounds (which have been used as selection heuristics in SAEAs \cite{liu_gaussian_2014}). We found that the confidence bounds have properties similar to the maximum and minimum, so we retained only the first 3 statistics for the rest of our experiments. At initialisation, the fitness of all infeasible solutions is set to a small, positive value $\epsilon$, causing the algorithm to select infeasible solutions uniformly. The surrogate model is trained as soon as one datapoint is available, and is updated online whenever new data is available. Whenever the model is updated, we use it to reassign fitnesses for recently-generated infeasible solutions.
    
    FI-2Pop has previously been combined with MAP-Elites to form CMAP-Elites \cite{khalifa_talakat_2018}, a quality-diversity algorithm that can find solutions that satisfy given constraints. In our current work, we demonstrate our novel variant of FI-2Pop within CMAP-Elites. The most benefit comes from using an optimising emitter (denoted \enquote{E}), which selects bins with the highest fitness per population. As the optimum choice of statistic and/or emitter may change per domain, or even over time, we use the $\epsilon$-greedy bandit algorithm \cite{sutton_reinforcement_1998} (denoted \enquote{B}) to select both in order to maximise fitness and coverage.

%% file: sections/05_results.tex
\section{Results}\label{sec:results}
    We demonstrate the performance of our proposed FI-2Pop variant over 3 experiments using the previously proposed hybrid EA in the Space Engineers domain \cite{gallotta_evolving_2022}: FI-2Pop, FI2-Pop within CMAP-Elites, and CMAP-Elites with emitters. For all experiments we ran each method for 50 generations, and provide statistics collected over 20 different random seeds.
    
\begin{figure}[ht!]
        \centering
        \begin{subfigure}{.24\textwidth}
            \centering
            \includegraphics[width=\linewidth]{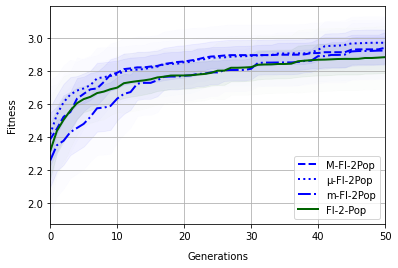}
            \caption{Elite feasible fitness}
            \label{fig:fi2pop-comparison-top-feas-fitnesses}
        \end{subfigure}%
        \begin{subfigure}{.24\textwidth}
            \centering
            \includegraphics[width=\linewidth]{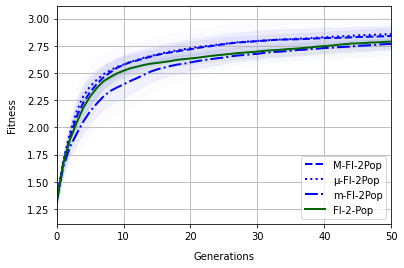}
            \caption{Avg. feasible fitness}
            \label{fig:fi2pop-comparison-avg-feas-fitnesses}
        \end{subfigure}
        \caption{FI-2Pop: elite and average fitness values for the feasible population, with FI-2Pop and our proposed variant with different statistics. Values show mean $\pm 1$ standard deviation.}
        \label{fig:fi2pop_comparisons}
    \end{figure}

    In our first experiment, we compared the standard FI-2Pop algorithm and our variant with different statistics. We show the feasible population's fitness over generations in \Cref{fig:fi2pop_comparisons}. All statistics result in a higher elite fitness, which is the final solution given by FI-2Pop. $\mu$-FI-2Pop gave the highest top and average fitnesses overall.
    In our second experiment, we tested CMAP-Elites with standard FI-2Pop EA and our variants. We set CMAP-Elites to operate on a fixed $32\times32$ grid, using axis ratios (see \Cref{sec:domain}) as the BCs.\footnote{We note that this choice of BCs restricts the maximum coverage to $50\%$.} Under this setting, we found that our variants perform similarly to CMAP-Elites.
    
\begin{figure}[ht!]
        \centering
        \begin{subfigure}{.24\textwidth}
            \centering
            \includegraphics[width=\linewidth]{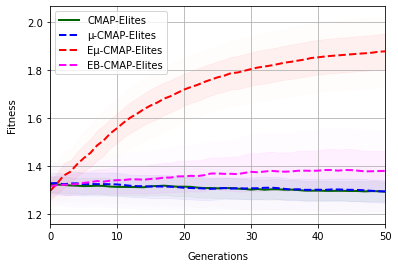}
            \caption{Avg. feasible fitness}
            \label{fig:mapelites-omparison-avg-feas-fitnesses}
        \end{subfigure}%
        \begin{subfigure}{.24\textwidth}
            \centering
            \includegraphics[width=\linewidth]{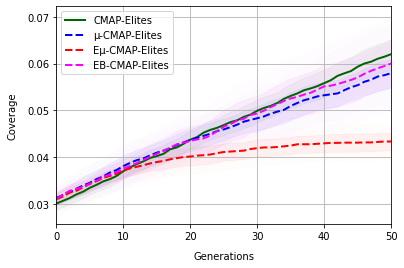}
            \caption{Avg. coverage}
            \label{fig:mapelites-comparison-avg-feas-fitnesses}
        \end{subfigure}
        \caption{CMAP-Elites: elite and average fitness values for the feasible population, with CMAP-Elites and our proposed variant with different statistics. Values show mean $\pm 1$ standard deviation.}
        \label{fig:mapelites_comparisons}
    \end{figure}

\begin{figure}[ht!]
        \centering
        \begin{subfigure}{.23\textwidth}
            \centering
            \includegraphics[width=\linewidth]{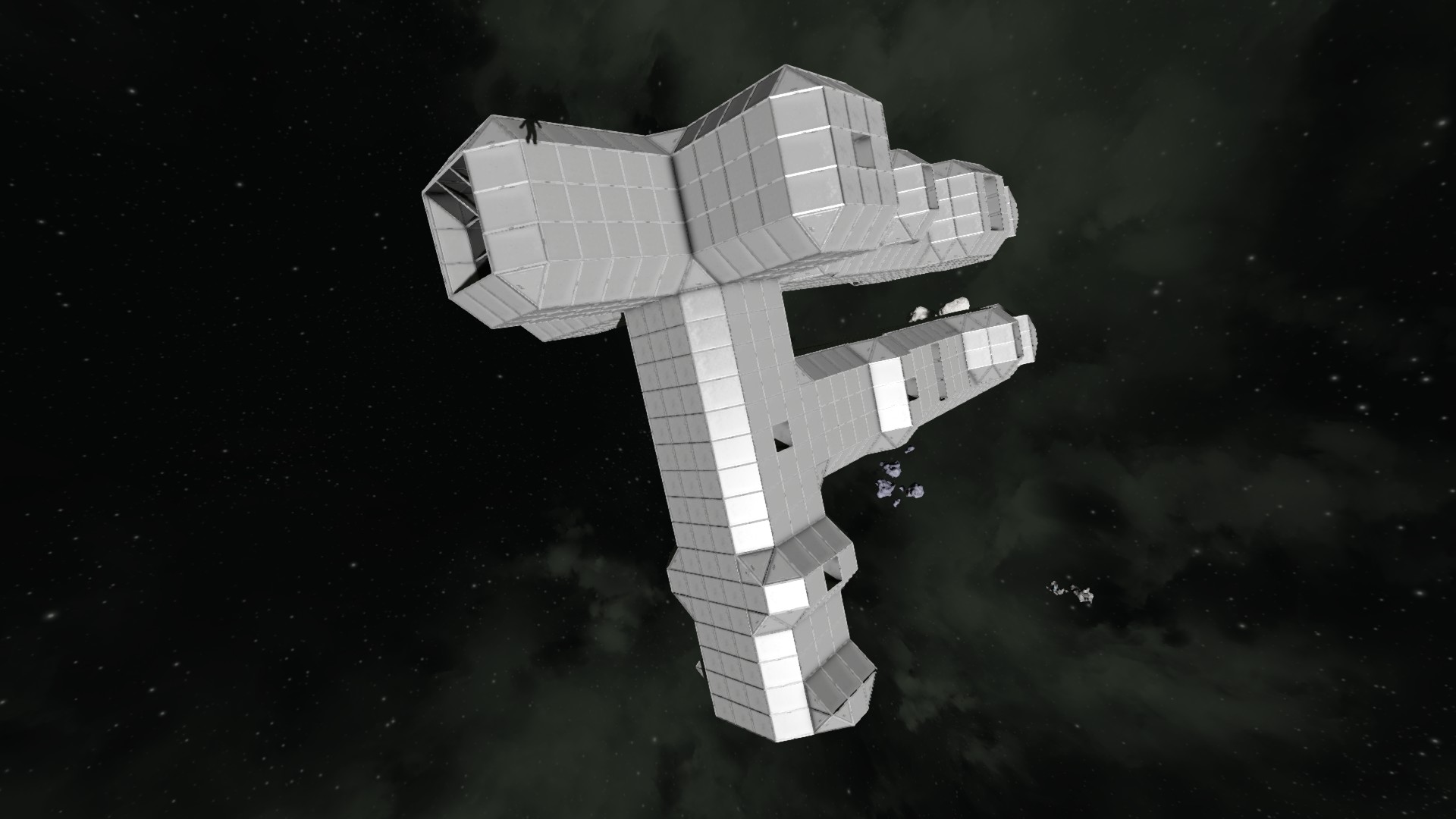}
            \caption{FI-2Pop elite content}
            \label{fig:fi2pop-content}
        \end{subfigure}%
        \begin{subfigure}{.23\textwidth}
            \centering
            \includegraphics[width=\linewidth]{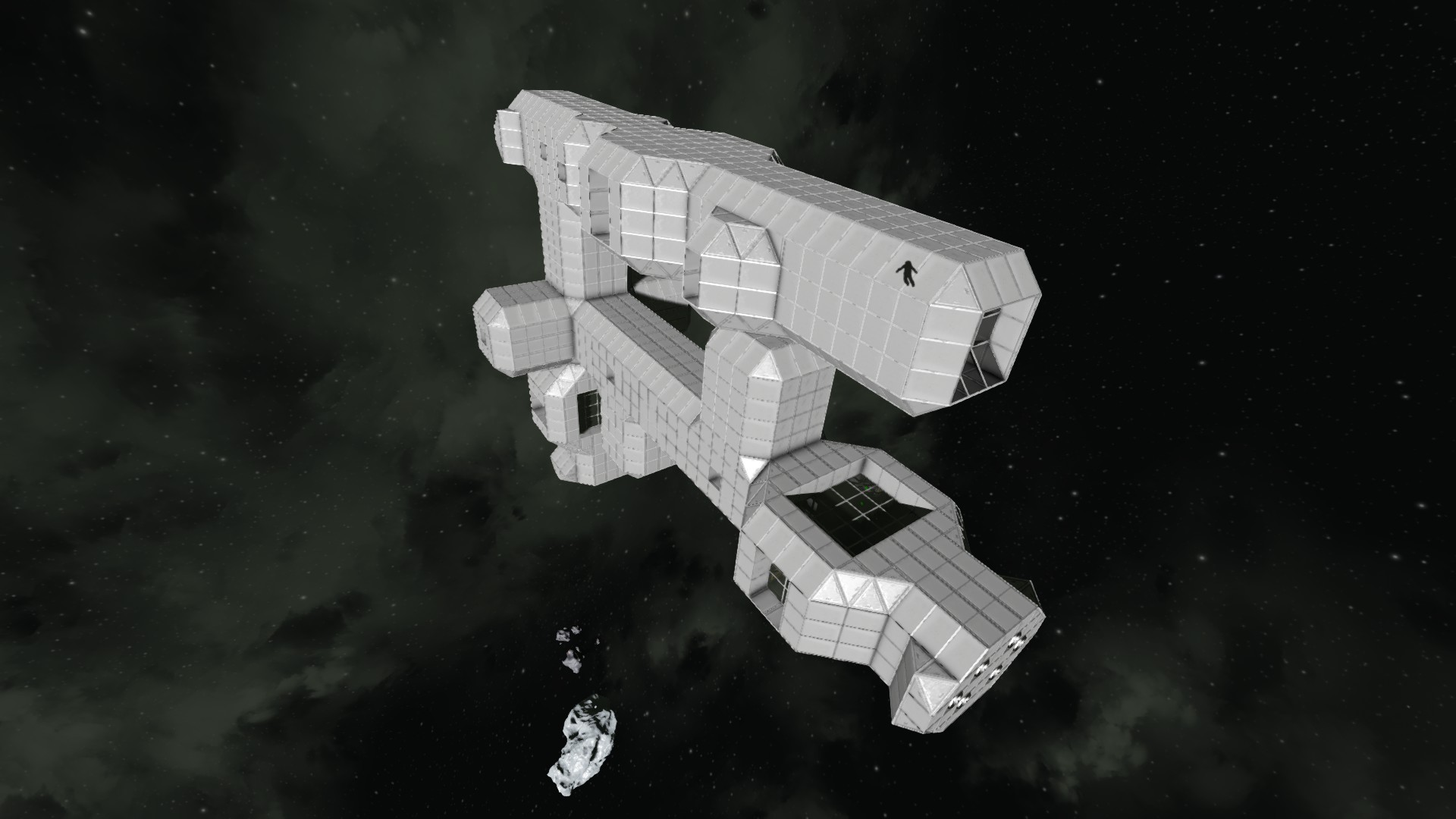}
            \caption{$\mu$-FI-2Pop elite content}
            \label{fig:mufi2pop-content}
        \end{subfigure}
        \begin{subfigure}{.23\textwidth}
            \centering
            \includegraphics[width=\linewidth]{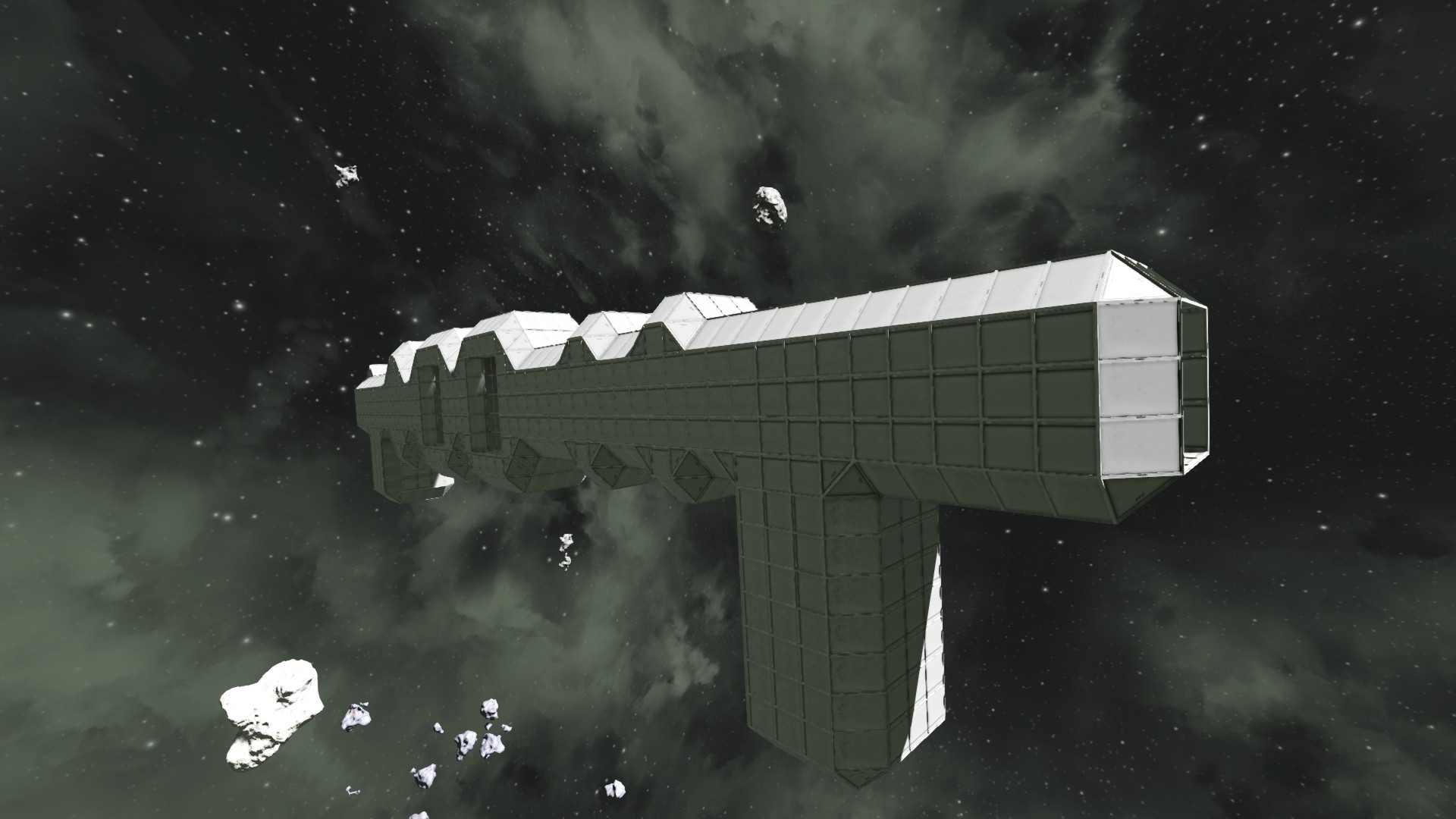}
            \caption{CMAP-Elites content (1)}
            \label{fig:mapelites-content_1}
        \end{subfigure}%
        \begin{subfigure}{.23\textwidth}
            \centering
            \includegraphics[width=\linewidth]{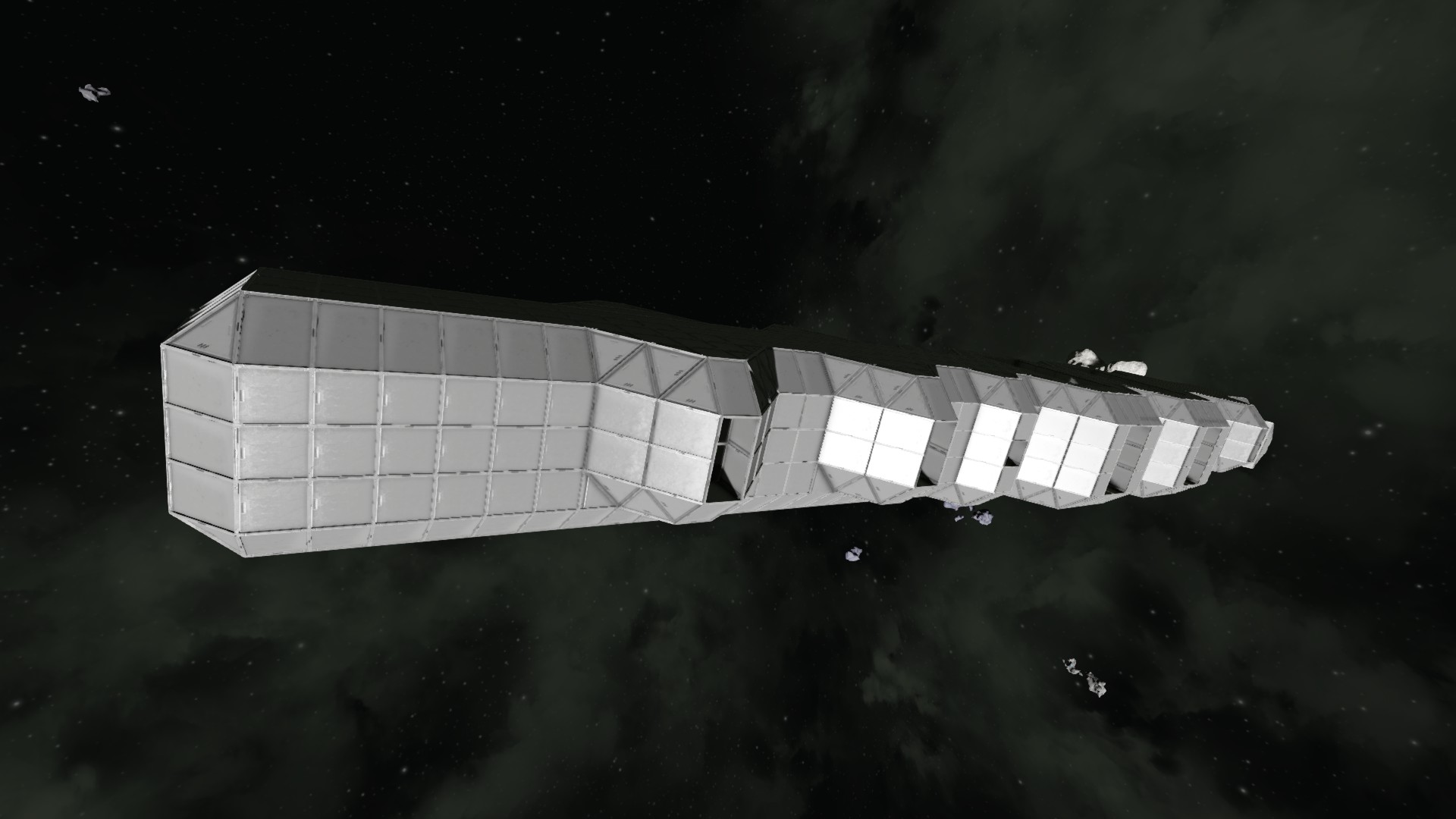}
            \caption{CMAP-Elites content (2)}
            \label{fig:mapelites-content_2}
        \end{subfigure}
        \begin{subfigure}{.23\textwidth}
            \centering
            \includegraphics[width=\linewidth]{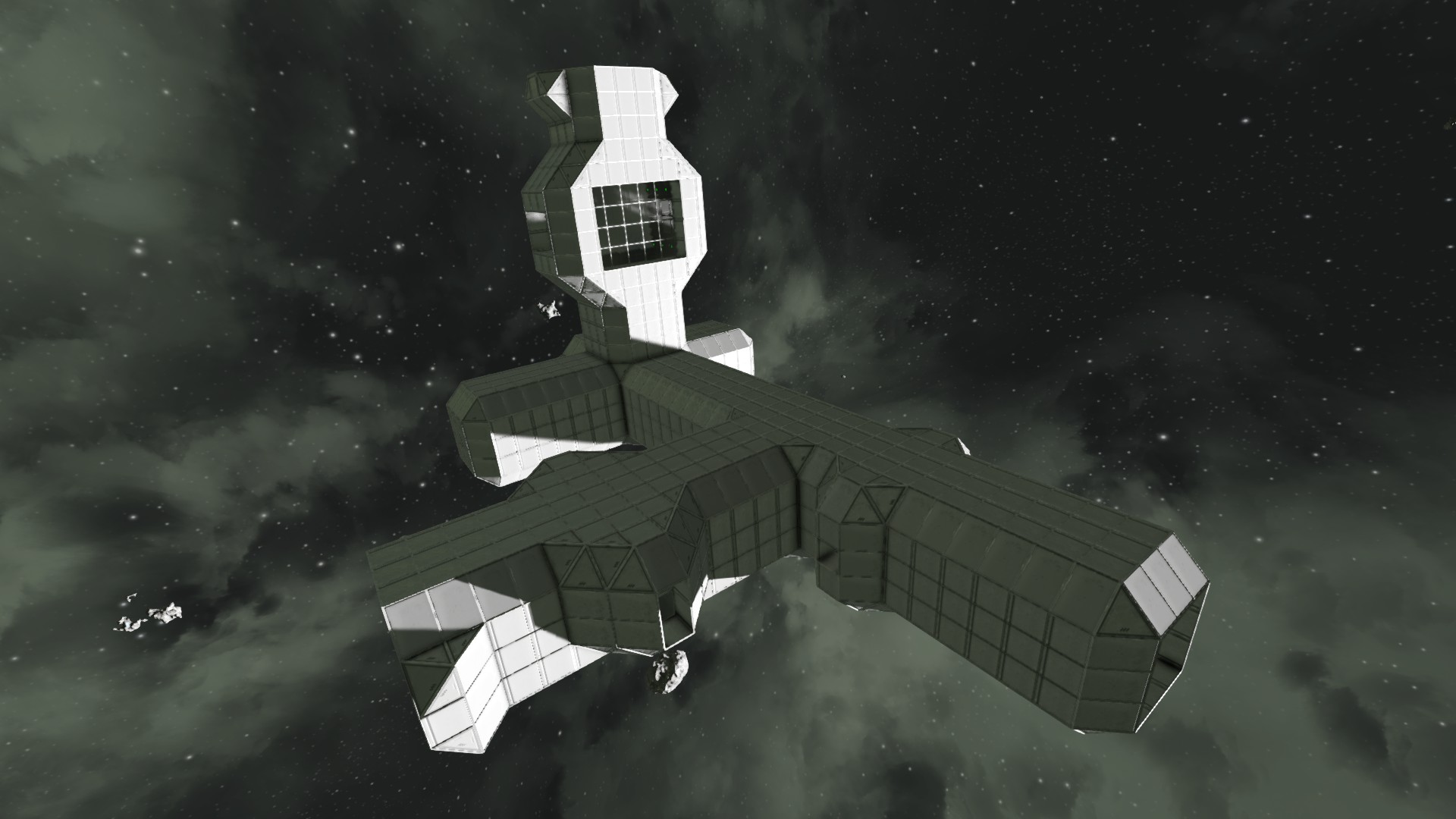}
            \caption{E$\mu$-CMAP-Elites content (1)}
            \label{fig:fmumapelites-content_1}
        \end{subfigure}%
        \begin{subfigure}{.23\textwidth}
            \centering
            \includegraphics[width=\linewidth]{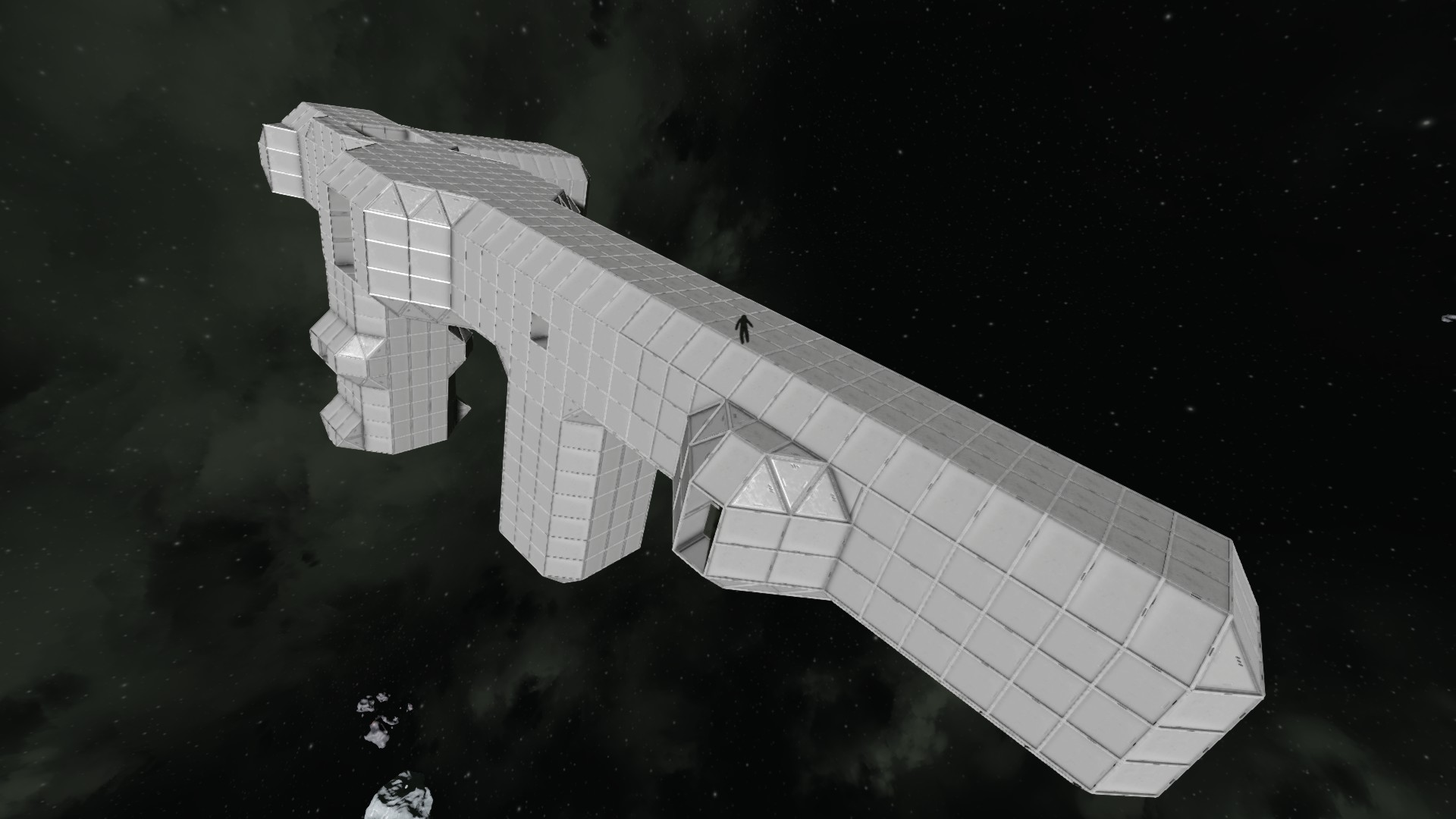}
            \caption{E$\mu$-CMAP-Elites content (2)}
            \label{fig:fmumapelites-content_2}
        \end{subfigure}
        \caption{In-game screenshots of elite solutions for different experiments, picked at random. The CMAP-Elites and E$\mu$-CMAP-Elites screenshots highlight elites from bins with different BCs (axis ratios).}
        \label{fig:ingame-contents}
    \end{figure}

    In our third experiment, we replaced the standard MAP-Elites selection rule (a \enquote{random emitter}) with the optimising emitter ($E$). We found that this resulted in a much higher feasible fitness overall, at the cost of a lower coverage. To compensate for this cost, we used an $\epsilon$-greedy bandit ($B$) to optimise the percentage increase of feasible fitness plus the percentage increase of coverage. This setup produced high-fitness solutions (higher than standard CMAP-Elites, lower than $E$) at a high coverage (comparable to standard CMAP-Elites, higher than $E$). We show the average fitness and coverage of the most salient CMAP-Elites experiments in \Cref{fig:mapelites_comparisons}.
    We report final metrics for all the different experiments in \Cref{tab:all_exp_metrics}. From these results, we demonstrate that our variant can find solutions at a higher fitness, as compared to the traditional FI-2Pop algorithm. A similar conclusion can be drawn for the application of our variant to CMAP-Elites, when combined with the optimising emitter. Finally, we note that using a bandit to select the emitter and statistic allows us to find an advantageous trade-off between feasible fitness and coverage, and can also be adapted to optimise different metrics, or combinations thereof.
    \begin{table}[]
        \centering
        \resizebox{.5\textwidth}{!}{%
            \begin{tabular}{|c|c|c|c|c|c|}
                \hline
                \multirow{2}{*}{Method} & \multicolumn{2}{c|}{Feas. fitness $\uparrow$} & \multicolumn{2}{c|}{Infeas. fitness $\uparrow$} & \multirow{2}{*}{Coverage $\uparrow$} \\
                \cline{2-5} & \multicolumn{1}{c|}{Top} & Avg. & \multicolumn{1}{c|}{Top} & Avg. & \\ \hline
                FI-2Pop               & $2.8856$ & $2.7882$ & -    & -    & - \\ \hline
                M-FI-2Pop           & $2.9426$ & $2.8402$ & $\mathbf{0.1270}$ & $\mathbf{0.1248}$ & - \\ \hline
                $\mu$-FI-2Pop       & $\mathbf{2.9732}$ & $\mathbf{2.8574}$ & $0.1146$ & $0.1118$ & - \\ \hline
                m-FI-2Pop           & $2.9259$ & $2.7673$ & $0.0645$ & $0.0638$ & - \\ \hline
                \hline \hline
                CMAP-Elites            & $2.5837$ & $1.2946$ & -    & -    & $0.062$ \\ \hline
                M-CMAP-Elites        & $2.5454$ & $12863$ & $0.1513$ & $0.0786$ & $0.0615$  \\ \hline
                $\mu$-CMAP-Elites       & $\mathbf{2.5850}$ & $\mathbf{1.2955}$ & $\mathbf{0.1723}$ & $\mathbf{0.0910}$ & $0.0580$  \\ \hline
                m-CMAP-Elites        & $2.4536$ & $1.2916$ & $0.0937$ & $0.0583$ & $\mathbf{0.0631}$  \\ \hline \hline \hline
                EM-CMAP-Elites  & $2.7163$ & $1.8547$ & $0.0293$ & $0.0126$ & $0.0427$  \\ \hline
                E$\mu$-CMAP-Elites & $\mathbf{2.7967}$ & $\mathbf{1.8774}$ & $0.0210$ & $0.0148$ & $0.0433$  \\ \hline
                Em-CMAP-Elites  & $2.7819$ & $1.8400$ & $\mathbf{0.0327}$ & $\mathbf{0.0249}$ & $0.0425$  \\ \hline
                EB-CMAP-Elites & $2.7529$ & $1.3808$ & N/A & N/A & $\mathbf{0.0600}$ \\ \hline
            \end{tabular}%
        }
        \caption{Average final fitness and coverage for all experiments. \textbf{Bold} values indicate the best result within each setting.}
        \label{tab:all_exp_metrics}
    \end{table}
    Finally, we give a qualitative overview of the spaceships generated by the different approaches. We found that our FI-2Pop variant is able to generate more complex solutions than those produced by standard FI-2Pop (see \Cref{fig:mufi2pop-content} compared to \Cref{fig:fi2pop-content}). Spaceships generated by standard CMAP-Elites are relatively simple (see \Cref{fig:mapelites-content_1,fig:mapelites-content_2}), and the use of our variant with the optimising emitter can find more interesting solutions (see \Cref{fig:fmumapelites-content_1,fig:fmumapelites-content_2}), while retaining comparable diversity if used in combination with the bandit algorithm.

%% file: sections/06_discussion.tex
\section{Discussions}\label{sec:discussions}
    In this work we presented a variant of FI-2Pop that uses a surrogate model to compute a different fitness for the infeasible population, based on the ability of a feasible solution to generate high-performing feasible solutions, which drives selection pressure towards feasible solutions that also have higher fitnesses. We also tested CMAP-Elites with both standard FI-2Pop and our variant. When combined with an optimising emitter, our method outperforms standard CMAP-Elites. Finally, we showed how the simple $\epsilon$-greedy bandit algorithm can automatically choose between emitters and metrics to achieve both high feasible fitness and coverage of the search space.
    
    We note that there are several considerations to using our method. Firstly, training the surrogate model online could be resource-intensive depending on the domain and choice of model. Secondly, our choice of fitness function performs better in domains where population hopping (in particular, when infeasible parents produce feasible offspring) is frequent, as more data is available to train the model. Finally, our method does not explicitly take into account the number of constraints violated, which may be a more important signal in other domains. Nevertheless, the concept of using a surrogate model to construct a more informative fitness function for FI-2Pop's infeasible population has potential benefits for PCG, and even wider applications requiring constrained optimisation.


%% file: sections/07_acks.tex
\section*{Acknowledgements}
\noindent This project was partly funded by a GoodAI research grant.